\documentclass[lettersize,journal]{IEEEtran}
\usepackage{amsmath,amsfonts}
\usepackage{algorithmic}
\usepackage{algorithm}
\usepackage{array}
\usepackage{textcomp}
\usepackage{stfloats}
\usepackage{url}
\usepackage{verbatim}
\usepackage{graphicx}
\usepackage{hyperref}
\usepackage{cleveref}
\usepackage{cite}
\usepackage{gensymb} 
\usepackage{caption}
\usepackage{subcaption}
\usepackage[table]{xcolor}
\usepackage{comment}
\usepackage{booktabs}
\usepackage{tabularx}
\usepackage{import}
\usepackage{bm}
\usepackage{siunitx}
\usepackage{acro}
\DeclareAcronym{com}{ 
    short = {CoM}, 
    long  = {Center of Mass},
    first-style = short-long,
}
\DeclareAcronym{cog}{ 
    short = {CoG}, 
    long  = {Center of Geometry},
    first-style = short-long,
}

\DeclareAcronym{wrt}{ 
    short = {w.r.t.}, 
    long  = {with respect to},
    first-style = short-long,
}
\DeclareAcronym{ee}{ 
    short = {EE}, 
    long  = {End-Effector},
    first-style = short-long,
}
\DeclareAcronym{dof}{ 
    short = {DoF}, 
    long  = {Degree of Freedom},
    first-style = short-long,
}
\DeclareAcronym{uav}{ 
    short = {UAV}, 
    long  = {Unmanned Aerial Vehicle},
    first-style = short-long,
}
\usepackage{csquotes}
\begin{document}

\title{AEROBULL: A Center-of-Mass Displacing Aerial Vehicle Enabling Efficient High-Force Interaction}
%\title{AEROBULL: A Center-of-Mass Displacing Aerial Vehicle Enabling Efficient High Force Pushing}

\author{ Tong Hui$^1$,~\IEEEmembership{Student Member,~IEEE,} Esteban Zamora$^1$, Simone D'Angelo$^2$,~\IEEEmembership{Student Member,~IEEE,}\\ Stefan Rucareanu$^1$,
 Matteo Fumagalli$^1$,~\IEEEmembership{Member,~IEEE}
\thanks{This work has been supported by the European Unions Horizon 2020 Research and Innovation Programme AERO-TRAIN under Grant Agreement No. 953454. Corresponding author email: {\tt\small tonhu@dtu.dk}}% <-this % stops a space
\thanks{${}^1$ Technical University of Denmark, Denmark}
\thanks{${}^2$ CREATE Consortium and PRISMA Lab, Department of Engineering and Information Technology, University of Naples Federico \uppercase\expandafter{\romannumeral2}}
}
%\thanks{Manuscript received April 19, 2021; revised August 16, 2021.}}

\markboth{Journal of \LaTeX\ Class Files,~Vol.~14, No.~8, August~2021}%
{Shell \MakeLowercase{\textit{et al.}}: A Sample Article Using IEEEtran.cls for IEEE Journals}

\IEEEpubid{0000--0000/00\$00.00~\copyright~2021 IEEE}

\maketitle

\begin{abstract}
In various industrial sectors, inspection and maintenance tasks using \ac{uav} require substantial force application to ensure effective adherence and stable contact, posing significant challenges to existing solutions.
This paper addresses these industrial needs by introducing a novel lightweight aerial platform ($3.12\si{\kilogram}$) designed to exert high pushing forces on non-horizontal surfaces. To increase maneuverability, the proposed platform incorporates tiltable rotors with 5-\ac{dof} actuation. Moreover, it has an innovative shifting-mass mechanism that dynamically adjusts the system's \ac{com} during contact-based task execution. A compliant \ac{ee} is applied to ensure a smooth interaction with the work surface. 
We provide a detailed study of the \ac{uav}'s overall system design, hardware integration of the developed physical prototype, and software architecture of the proposed control algorithm. Physical experiments were conducted to validate the control design and explore the force generation capability of the designed platform via a pushing task. With a total mass of $3.12\si{\kilogram}$, the \ac{uav} exerted a maximum pushing force of above $28\si{\newton}$ being almost equal to its gravity force. Furthermore, the experiments illustrated the benefits of having displaced \ac{com} by benchmarking with a fixed \ac{com} configuration.
\end{abstract}


\section{introduction}\label{s1:intro}
Inspection and maintenance tasks on non-horizontal surfaces in industrial settings often require substantial force exertion and precise control to maintain stable contact with surfaces. Traditional manual inspections using human workers face limitations in accessing hard-to-reach areas and
% maintaining stability, and managing the high contact forces necessary for effective operations. These limitations can compromise operational efficiency and safety. Moreover, critical working conditions in industrial applications, such as oxygen-deprived tunnels, and offshore environments,
often lead to issues related to workspace safety, a shortage of skilled labor, and high costs~\cite{construct}. 
\begin{figure}
    \centering
    \includegraphics[width=\columnwidth]{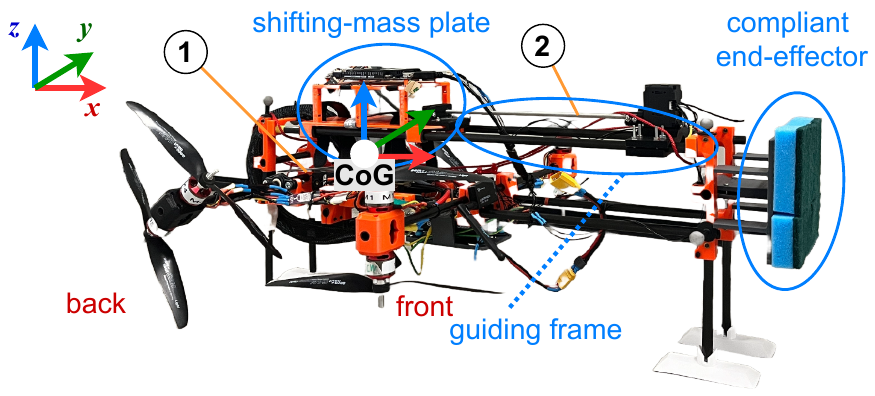}\\
    \includegraphics[width=\columnwidth]{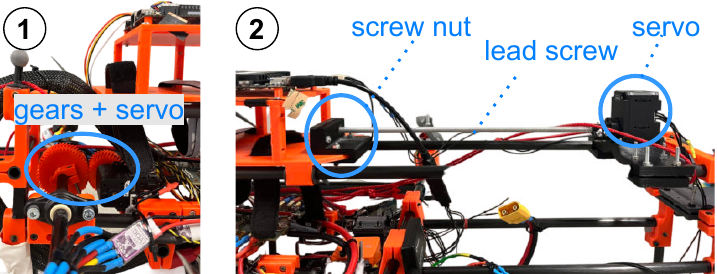}
    \caption{Designed aerial platform with tiltable back rotors, the shifting-mass plate, and the guiding frame. Zoom for the back rotors' tilting mechanism (1) and the \ac{com} displacing mechanism (2). The body frame $\mathcal{F}_B$'s origin coincides with the system's \ac{cog}.}
    \label{fig:real_platform}
\end{figure}
Recently, the use of \ac{uav}s, especially multirotor platforms, that can physically interact with the environment for such tasks has gained attention. The development of these platforms has advanced considerably over the past decade~\cite{review_0,review}.

\begin{figure}[t]
   \centering
\begin{subfigure}{0.49\linewidth}\centering
    \includegraphics[trim={0.5cm 0.4cm 0.7cm  0.7cm},clip,width=\columnwidth]{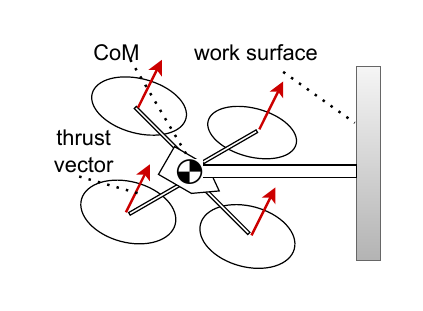}
    \caption{}
    \label{fig:unidirection}
\end{subfigure}
    \begin{subfigure}{0.47\linewidth}\centering
    \includegraphics[trim={0.5cm 0.4cm 0.7cm  0.6cm},clip,width=\columnwidth]{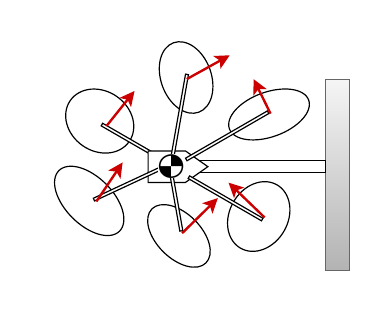}
    \caption{}
    \label{fig:multidirection}
\end{subfigure}\\
\:\:\:\;\;\;
\begin{subfigure}{0.44\linewidth}\centering
    \includegraphics[trim={0.0cm 0.0cm 0.0cm  0.0cm},clip,width=\columnwidth]{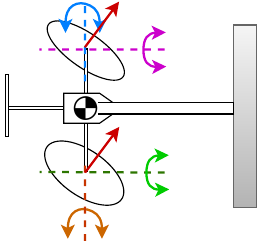}
    \caption{}
    \label{fig:thrustvector_2}
\end{subfigure}
\begin{subfigure}{0.42\linewidth}\centering
    \includegraphics[trim={0.0cm 0.0cm 0.0cm  0.0cm},clip,width=\columnwidth]{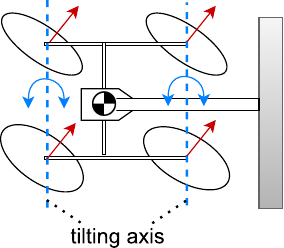}
    \caption{}
    \label{fig:thrustvector_1}
\end{subfigure}
\caption{Existing multirotor platforms interacting with vertical surfaces with a focus on the rotor configuration and the system's \ac{com} location during the pushing task execution.}\label{fig:related_work}
    %\vspace{-0.5cm}
\end{figure}

Various types of interaction tasks require the aerial system to exert different forces and torques in terms of both direction and magnitude~\cite{tong_aim}. The conventional multirotors, e.g., quadrotors, and hexacopters, feature a simple design and unidirectional thrust generation \ac{wrt} the \ac{uav} body. The inherent underactuated nature of such systems presents coupled gravity compensation and force generation on non-horizontal surfaces~\cite{ding,icra_tong_dtu,park}. To allow force exertion and stable contact on non-horizontal surfaces using these platforms, higher \ac{dof} manipulators are often mounted on the \ac{uav}~\cite{kim,dangeloRAS,fabio,bart}. Considering a pushing task on a vertical surface with only a rigid link \cite{tong_aim,icra_tong_dtu}, the \ac{uav} has to tilt its whole body to apply a pushing force normal to the surface as in \cref{fig:unidirection}.

\IEEEpubidadjcol

Alternatively, aerial systems with thrust vectoring capability allow multidirectional interactions with non-horizontal surfaces with only a rigidly attached link, see \cref{fig:multidirection}, \cref{fig:thrustvector_2}, and \cref{fig:thrustvector_1}. This can be achieved by using either fixed-titled rotors~\cite{park,tognon2018} or tiltable rotors~\cite{ding,bodie2019,mina,truj,watson2022,hwang,icra_tong_asl}. In~\cite{park,tognon2018,bodie2019,mina,truj,icra_tong_asl}, authors introduced fully-actuated aerial vehicles with decoupled linear and angular motion. These platforms can exert 6-\ac{dof} forces and torques in arbitrary directions (omnidirectional), being desirable for physical interactions with work surfaces at various orientations, as illustrated in \cref{fig:multidirection}. However, these rotor configuration designs often suffer from counteracting unwanted thrust components, which are inefficient. In \cite{watson2022}, authors present the Voliro Tricopter aerial manipulation platform with two pairs of bi-axial tiltable rotors as in \cref{fig:thrustvector_2}, while still being fully-actuated. In \cite{ding,hwang}, authors present underactuated aerial manipulation platforms with 5-\ac{dof}, being more efficient during free-flights and manipulation tasks like pushing and drilling, as in \cref{fig:thrustvector_1}.

\begin{figure*}[b]
   \centering
   \begin{subfigure}{0.17\linewidth}\centering
    \includegraphics[trim={0.9cm 0.6cm 0.7cm  0.7cm},clip,width=\columnwidth]{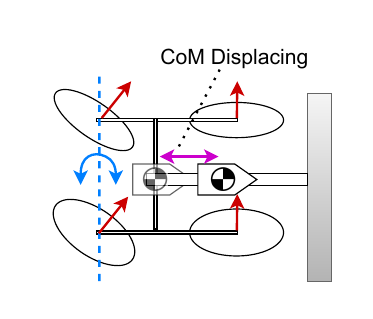}
    \caption{}
    \label{fig:com_displace}
\end{subfigure}
\begin{subfigure}{0.27\linewidth}\centering
    \includegraphics[trim={0.9cm 0.4cm 0.9cm  0.4cm},clip,width=\columnwidth]{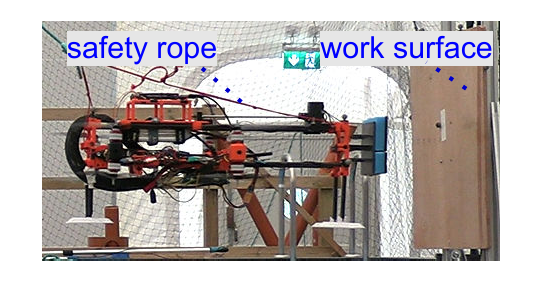}
    \caption{}
    \label{fig:action_1}
\end{subfigure}
    \begin{subfigure}{0.27\linewidth}\centering
    \includegraphics[trim={0.9cm 0.4cm 0.9cm  0.4cm},clip,width=\columnwidth]{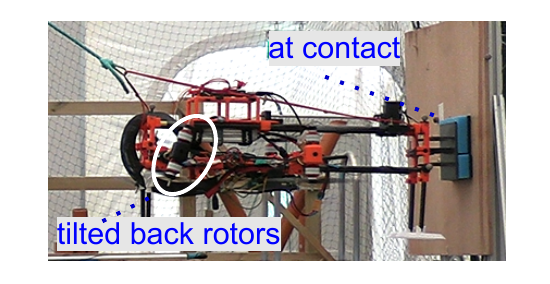}
    \caption{}
    \label{fig:action_2}
\end{subfigure}
\begin{subfigure}{0.27\linewidth}\centering
    \includegraphics[trim={0.9cm 0.7cm 0.9cm  0.4cm},clip,width=\columnwidth]{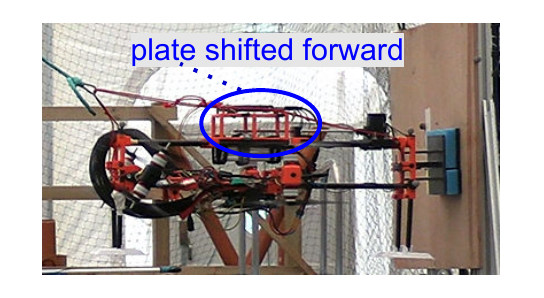}
    \caption{}
    \label{fig:action_3}
\end{subfigure}
\caption{(a) Proposed system design. (b-d) Snapshots from the physical experiment illustrate the system's operational flow for a pushing task on vertical surfaces. (b) The aerial vehicle follows the way-points to approach the work surface. (c) The platform establishes a contact using the compliant \ac{ee} while the back rotors are tilted to exert horizontal forces normal to the surface. (d) The shifting-mass plate is shifted toward the work surface to exert higher pushing forces.}\label{fig:actions}
    %\vspace{-0.5cm}
\end{figure*}
On the other hand, ensuring substantial force generation normal to the work surface is pivotal in contact-based inspection and maintenance tasks \cite{hwang,drill,sun,grinding}. Different approaches have been developed to allow the aerial system to exert high forces \ac{wrt} the system mass. A rather straightforward method is to use co-axial rotors with higher thrust generation together with decoupled gravity compensation and force generation on the surface \cite{bodie2019,hwang}. The existing multi-rotor platforms, however, generally have a fixed system \ac{com} within the rotor-defined area during both free-flights and physical interactions, as in \cref{fig:related_work}. During interactions, such system design with a fixed \ac{com} leads to a large momentum arm between the \ac{ee} tip and the \ac{com}. This configuration limits the system's force exertion on work surfaces due to noticeable torques introduced by the large momentum arm.
In \cite{drill},
the authors introduced a perching and tilting aerial vehicle
designed for concrete wall drilling. This aerial vehicle has two
suction cups for perching and can bring its fixed \ac{com} closer
to the work surface by maneuvering the aerial vehicle. During
these operations, ensuring successful perching between the
aerial platform and the work surface is essential. However, industrial infrastructures often present complex environments that can affect perching robustness, leading to risky conditions \cite{meng}.
\subsection{Contribution}
To enhance aerial manipulation technologies in efficient high-force interaction on non-horizontal surfaces, in our previous work \cite{ral_aerobull} we proposed the detailed modeling and control of a novel aerial platform with \ac{com} displacing capability and tiltable back rotors, see \cref{fig:real_platform}. The tiltable back rotors allow the platform to interact with non-horizontal surfaces with only a rigidly attached link. Moreover, we introduce an innovative shifting-mass mechanism allowing for dynamical displacement of the \ac{com} toward the work surface during interactions. The system presents the configuration shown in \cref{fig:com_displace} and an operational flow displayed in \cref{fig:action_1,fig:action_2,fig:action_3}. Such system design allows the platform to exert higher forces on the vertical surface by only tilting back rotors, being more energy efficient \ac{wrt} the configurations in \cref{fig:multidirection,fig:thrustvector_2,fig:thrustvector_1}. 

In addition to our previous work \cite{ral_aerobull}, with this article, we present:
\begin{enumerate}
    \item the detailed hardware design of the built physical prototype including both mechanical design and mechatronic components;
    \item the software development and integration for the proposed control design;
    \item additional physical experiments illustrating the high-force generation capability through a pushing task on a vertical surface highlighting the developed prototype's key aspects and performances.
\end{enumerate}
% This mechanism is vital for increasing pushing forces on vertical surfaces and maintaining flight stability. Additionally, our drone features a compliant \ac{ee} that ensures smooth interaction with the work surface, increasing friction and stability during operations.
% In this paper, we provide a comprehensive analysis of the drone's mechanical and electronic configuration, as well as the developed control system. We detail how the shifting-mass mechanism facilitates precise \ac{com} adjustments, enabling the drone to exert and manage greater contact forces effectively. Experimental results demonstrate the capability of our drone to engage safely and efficiently with vertical surfaces, showcasing improved performance and stabilization compared to existing models.
\subsection{Paper Outline}
The remaining part of the article is outlined as the following:
\begin{itemize}
    \item \cref{s2:sys_des} and \cref{sec:hardware}: the overall aerial vehicle system design and detailed hardware integration;
    \item \cref{s3:sys_model} and \cref{s4:ctrl_des}: aerial vehicle system modeling and the developed control pipeline;
    \item \cref{s6:sftw}: implemented software architecture;
    \item \cref{s7:case_st}: indoor contact-based physical experiments with the developed aerial vehicle;
    \item \cref{s8:end}: conclusions.
\end{itemize}

\section{Overall System Design} \label{s2:sys_des}
\begin{figure}[t]
   \centering
\includegraphics[width=\columnwidth]{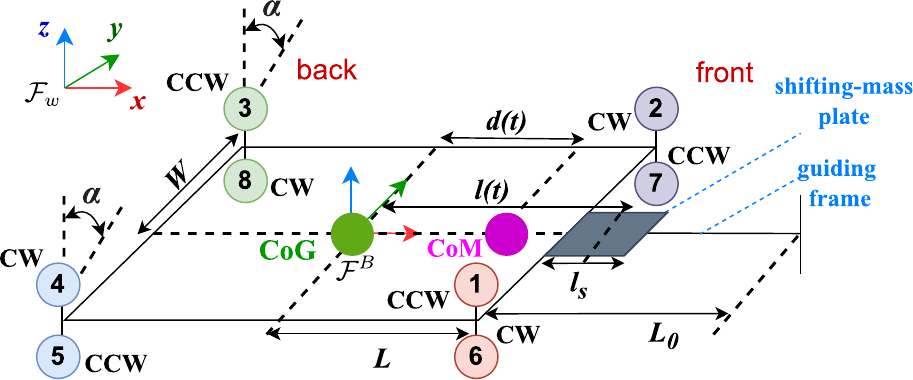}
    \caption{The world frame $\mathcal{F}_w$ and the body frame $\mathcal{F}^B$ are depicted. Rotors 3, 4, 5, and 8 can tilt simultaneously with $\alpha \in [-90\degree, \ 90\degree]$ around $\bm{y}_B$; the system's \ac{com} can displace along $\bm{x}_B$ via changing the shifting-mass plate position. CCW: counter-clockwise; CW: clockwise.}
    \label{fig:system_modeling}
    %\vspace{-0.5cm}
\end{figure}
\begin{table}[t!]
\centering
\caption{Aerobull Physical Properties.}
\begin{tabularx}{\columnwidth}{p{0.15\textwidth}|p{0.15\textwidth}|X}
\toprule
\textbf{Property} & \textbf{Value} & \textbf{Unit} \\ \hline
$L$ & $0.138$ & \si{\meter} \\ 
$W$ & $0.225$ & \si{\meter} \\ 
$L_0$ & $0.180$ & \si{\meter}\\ 
$l_S$ & $0.200$ & \si{\meter}\\ 
$m$ & $3.12$ & \si{\kilogram} \\ 
$m_S$ & $0.90$ & \si{\kilogram} \\
\hline
\end{tabularx}
\label{table_param}
\end{table}
This section overviews the system design and the proposed aerial vehicle's operational flow.
The aerial platform adopts the form of an H-shaped coaxial octocopter with the body frame $\mathcal{F}^B=\{O_B;\bm{x}_B,\bm{y}_B,\bm{z}_B\}$ attached to its \ac{cog}, as shown in \cref{fig:real_platform}. Concerning the simplified schematic of the system model in~\cref{fig:system_modeling}, the front rotors ($1$, $2$, $6$, and $7$) have fixed rotating axes parallel to $\bm{z}_B$ while the back rotors ($3$, $4$, $5$, and $8$) have tiltable axes that can pivot around $\bm{y}_B$ simultaneously. $\alpha \in [-90\degree, \ 90\degree]$ denotes the tilting angle between each tiltable axis and $\bm{z}_B$. This design introduces an additional \ac{dof} compared to classic aerial vehicles with 4-\ac{dof} actuation. The distance from the system's \ac{cog} to each rotor center along $\bm{x}_B$ is represented by $L$, while the distance along $\bm{y}_B$ is denoted as $W$. 

Above all, the platform also features a displaceable \ac{com} along the body axis $\bm{x}_B$, defined as the interaction axis. This feature is enabled via sliding a plate equipped with heavy components along a \textit{guiding frame} parallel to $\bm{x}_B$, which we call the \textit{shifting-mass plate}, see \cref{fig:real_platform}. The length of the guiding frame is denoted as $L_0$, see \cref{fig:system_modeling}. An \ac{ee} is mounted at the tip of the guiding frame to contact the work surface. During a pushing task, firstly, the designed platform establishes stable contact between the \ac{ee} and the work surface, as in \cref{fig:action_2}. Once stable contact is ensured, the shifting-mass plate is moved toward the work surface along $\bm{x}_B$ to enable higher force exertion, as in \cref{fig:action_3}.

We use the \ac{cog} of the shifting-mass plate to present its location. We denote $l_S$ as the length of the shifting-mass plate along the body axis $\bm{x}_B$, as in \cref{fig:system_modeling}. The shifting-mass plate position along $\bm{x}_B$ \ac{wrt} the body frame origin $\bm{O}_B$ is denoted as $l(t) \in [0,L+L_0-0.5l_S]$. By changing the shifting-mass plate position, the resultant displacement of the system's \ac{com} along $\bm{x}_B$ is denoted by $d(t)$. When the system's \ac{com} is located at $d>L$ outside the rotor-defined area (i.e., over-displaced), the aerial vehicle often flips around the contact area, leading to instability. Therefore, $d(t)$ is restricted by $[0, L]$ to avoid risky scenarios and damage to the platform. Assuming the symmetric mass distribution of the platform around its body axes, the relation between $l(t)$ and $d(t)$ is given by: 
\begin{equation}\label{eq:shift_com}
    d(t)=\frac{m_S}{m}l(t).
\end{equation}
where $m_S$ denotes the mass of the shifting-mass plate while $m$ is the system total mass, with $m_S<m$. The maximum \ac{com} displacement $d=L$ results in a shifting-mass plate position of $l=\displaystyle\tfrac{m}{m_S}L>L$. 
\section{Hardware Prototype}\label{sec:hardware}
\begin{figure}[t!]
   \centering
\includegraphics[trim={0.5cm 0.0cm 0.0cm  0.0cm},clip,width=\columnwidth]{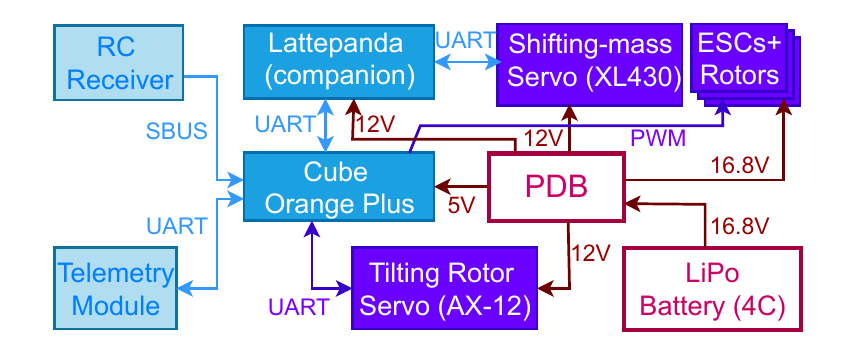}
    \caption{Hardware integration scheme: the main connections are depicted.}
    \label{fig:hardware}
    %\vspace{-0.5cm}
\end{figure}
\begin{table}[t!]
\centering
\caption{Aerobull Physical Components.}
\begin{tabularx}{\columnwidth}{p{0.13\textwidth}|p{0.22\textwidth}|X}
\toprule
\textbf{Component} & \textbf{Product Name}      &   \textbf{Quantity}                                                   \\ \hline
PDB                       & SkyDrones SmartAP &1                               \\ 
Battery                   & {Tattu 6750mAh 14.8V 25C 4S1P} & 1 \\ 
ESC                      & Flycolor Raptor5 ESC 45A &8                          \\ 
Motors                     & Turnigy D2826      &8                                 \\ 
Propellers                     & Aeronaut CAM Prop $9\times5$''         &8                    \\ 
Telemetry              & SiK Telemetry Radio V3   &1                           \\ 
RC receiver               & FrSky RX8R          &1                               \\ 
RC transmitter            & Taranis X9D         &1                                \\ 
Flight Controller         & Cubepilot Cube Orange Plus              &1          \\ 
Onboard Computer         & Lattepanda 3 Delta SBC                    &1          \\ 
Tilt-angle servo    & Dynamixel AX-12A                             &1       \\ 
Shifting-mass servo & Dynamixel XL430-W250                      &1          \\ \hline
\end{tabularx}
\label{table_components}
\end{table}
A physical prototype as in~\cref{fig:real_platform}, whose main physical properties are shown in~\cref{table_param}, is built from scratch using 3D printed parts, carbon fiber tubes, relying on the system design in~\cref{s2:sys_des}.
Its mechatronic devices are listed in~\cref{table_components}. In the following, we detail the hardware of the built physical prototype.
\subsection{Propulsion System}
The propulsion system includes the tiltable back rotors and the fixed front rotors. To tilt the rotating axes of the back rotors simultaneously around the body axis $\bm{y}_B$, the rotor pairs 4\&5 and 3\&8 are rigidly attached to a carbon fiber tube. This tube can be rotated via a pair of gears and a Dynamixel AX-12A servo motor as shown in the bottom left figure of \cref{fig:real_platform}. This smart motor supports a total range of $[0\degree, \ 300\degree]$ suitable for the allowed tilting angle range. Each rotor is composed of a FLYCOLOR Raptor5 Brushless ESC (Electronic Speed Controller) (3-6S, 45A), a Turnigy D2826 motor, and an Aeronaut CAM Propeller. We define $k_t$ and $k_b$ as each rotor's thrust and drag torque coefficients. An RC-Benchmark Series 1580 Test Stand was used to identify the coefficients of the designed propulsion system, with $k_t=1.156 \cdot 10^{-5}$ and $k_b=0.0277\cdot k_t$. The system achieves the total thrust required for gravity compensation with a reasonable speed ratio of $ \approx 0.5$ \ac{wrt} the maximum speed.
\subsection{\ac{com} Displacing}
As mentioned in \cref{s2:sys_des}, the system's \ac{com} displacing is achieved by changing the shifting-mass plate position along the guiding frame. This sliding motion of the shifting-mass plate is enabled by a linear actuator composed of a Dynamixel XL430 servo mounted at the guiding frame tip, a M$6$ lead screw, and a M$6$ screw nut inset in a box rigidly attached to the shifting-mass plate, illustrated in the bottom right figure of \cref{fig:real_platform}. This design ensures accurate, smooth, and slow motion of the dynamical \ac{com} displacing process. The selected servo motor supports the extended position (multi-turn) mode, which allows continuous mapping between the servo position and the shifting-mass plate position $l$.
\subsection{System Integration}
A compliant \ac{ee} is mounted at the tip of the guiding frame to reduce the impact during initial contact with the work surface, as in \cref{fig:real_platform}. A high static friction coefficient material is used at the \ac{ee} tip to provide sufficient friction for stable pushing. The Cubepilot Cube Orange Plus is selected as the flight controller with the Lattepanda 3 Delta as the onboard companion computer. The hardware connectivity diagram, detailing power distribution and communication interfaces is illustrated in~\cref{fig:hardware}. The Sky-Drones SmartAP PDB (Power Distribution Board) distributes the power from the 4C LiPo battery to the aerial vehicle's different components.
The flight controller outputs PWM (Pulse-Width Modulation) signals to command the ESC to control the motor power output and rotating speed. The telemetry radio facilitates remote configuration and monitoring through \texttt{QGroundControl}, while the RC (Radio-Controller) system handles manual control and safety failsafe functionalities. 
\section{system modeling}\label{s3:sys_model}
This section presents the modeling of the examined system: the main reference frames are depicted in \cref{fig:system_modeling} illustrating the~\cref{s2:sys_des} aerial vehicle's schematic. The body frame $\mathcal{F}^B$ orientation \ac{wrt} the world system ($\mathcal{F}_w=\{\bm{O};\bm{x},\bm{y},\bm{z}\}$) is expressed in the special orthogonal group $SO(3)$, through the rotation matrix $\bm{R} \in SO(3)$ of the roll-pitch-yaw Euler angles ($\phi$, $\theta$, $\psi \in (-90\degree,90\degree)$).

According to the rotor configuration in \cref{fig:system_modeling}, the thrust magnitude and the drag torque of the $i$-th rotor are given by $T_i=k_t\Omega_i^2 \in \mathbb{R}$ and $\tau_i =(-1)^i k_b \Omega_i^2 \in \mathbb{R}$, where $\Omega_i \in \mathbb{R}^+$ is the rotor's rotating speed.
To retrieve the system's dynamic model we neglect the moving plate dynamics assuming slow motion of the shifting-mass plate. The body frame $\mathcal{F}_B$ system equations of motion can be written as:
\begin{equation}
         \begin{bmatrix} \bm{M} & \bm{0} \\ \bm{0} &\bm{I}\end{bmatrix} \begin{bmatrix}
             \dot{\bm{\upsilon}} \\\dot{\bm{\omega}}  
         \end{bmatrix}+\begin{bmatrix}
             \mathbb{S}(\bm{\omega}) \bm{M}\bm{\upsilon} \\  \mathbb{S}(\bm{\omega}) \bm{I}\bm{\omega}  \end{bmatrix}+\bm{G}=\begin{bmatrix}
                 \bm{F}_a \\ \bm{\Gamma}_a
             \end{bmatrix}+ \begin{bmatrix}
                 \bm{F}_C \\ \bm{\Gamma}_C
             \end{bmatrix}.
             \label{eq:dyn_model}
\end{equation}
where $\bm{\upsilon},\bm{\omega} \in \mathbb{R}^{3}$ are the \ac{uav}'s linear and angular velocity vectors while $\bm{\dot{\upsilon}},\bm{\dot{\omega}} \in \mathbb{R}^{3}$ are their time derivatives; $\begin{bmatrix} \bm{F}_a \\ \bm{\Gamma}_a \end{bmatrix} \in \mathbb{R}^{6}$ is the stacked vector of the system actuation wrenches, and $\bm{M}, \bm{I} \in \mathbb{R}^{3\times 3}$ are the mass and inertia matrices. $\bm{G} \in \mathbb{R}^{6}$ is the gravity term expressed in the body frame. Considering the \ac{com} displacing, we have:
\begin{equation}
    \resizebox{0.9\columnwidth}{!}{$\bm{G}=\begin{bmatrix} \bm{R}^{\top}g\bm{e}_3\\\mathbb{S} (\bm{R}^{\top}g\bm{e}_3)\begin{bmatrix}
        -d(t)&0&0
    \end{bmatrix}^{\top}\end{bmatrix}=\begin{bmatrix}
        \bm{G}_{lin} \\ \bm{G}_{ang}
    \end{bmatrix}=\bm{G}(l)$},
\end{equation}
with $g=-9.81$\si{\meter\per\square\second}, $\bm{e}_3=\begin{bmatrix}
    0&0&1
\end{bmatrix}^{\top}$, and $d=\frac{m_S}{m}l$ from \cref{eq:shift_com}.
Assuming that the interaction wrenches at the end-effector tip are the only sources of external wrenches, we denote $\begin{bmatrix} \bm{F}_C \\ \bm{\Gamma}_C \end{bmatrix} \in \mathbb{R}^{6}$ as the stacked force and torque vector exerted from the environment to the system. In~\cref{eq:dyn_model}, we use $\mathbb{S}(\bm{a}) \in \mathbb{R}^{3\times3}$ to present the skew symmetric matrix such that $\mathbb{S}(\bm{a})\bm{b}=\bm{a}\times \bm{b}$, and we have $\dot{\bm{R}}=\bm{\omega}\times\bm{R}=\mathbb{S}(\bm{\omega})\bm{R}$. We define the operator $(\cdot)^{\vee}$ such that $\bm{a}=\mathbb{S}(\bm{a})^{\vee}$.
\subsection{Moment of Inertia Estimation} \label{sec:inertia}
Differently to conventional aerial vehicles, the shifting-mass mechanism in~\cref{fig:system_modeling} affects the system's inertia by varying the shifting-mass plate position $l(t)$ along the body axis $\bm{x}_B$\footnote{For simplicity, $\mathcal{F}_B$ body axes correspond with the system's principal axes of inertia.}. This displacement leads to modifications in the moments of inertia along the body axes $\bm{y}_B$ and $\bm{z}_B$. %From the practice, we noticed that $I_{xx}=0.0444$\si{\kilo\gram\square\meter} remains constant by varying $l$, while $I_{yy}$ and $I_{zz}$ change values along with $l$.
Starting from the realistic CAD models representing the platform in~\cref{fig:real_platform}, we employ linear regression technology to obtain mathematical models of $I_{yy}$ and $I_{zz}$ \ac{wrt} $l$ similarly to \cite{aero_tong}:
\begin{equation}\label{eq:inertia}
\begin{split}
    &I_{yy}(l)=0.49l^2+0.0538 \ (\si{\kilo\gram\square\meter}),\\
    &I_{zz}(l)=0.52l^2+0.0795 \ (\si{\kilo\gram\square\meter}),
    \end{split}
\end{equation}
while $I_{xx}=0.0444$\si{\kilo\gram\square\meter} remains constant.
The proposed system's inertia matrix is thus given by $\bm{I}(l)=diag(\begin{bmatrix}
    I_{xx}&I_{yy}(l)&I_{zz}(l)
\end{bmatrix})$.
\subsection{System Actuation}\label{sec:actuation}
To complete the formulation in~\cref{eq:dyn_model}, the vector\footnote{$C_{(\cdot)}$ and $S_{(\cdot)}$ denote $\cos(\cdot)$ and $\sin(\cdot)$, while $(\cdot)_{1,2,3,...,n}$ represents the sum of thrust and drag torques, e.g., $T_{3,4,5,8}=T_3+T_4+T_5+T_8$.\label{fn:repeat}} $\bm{\Gamma}_a \in \mathbb{R}^{3}$ \ac{wrt} the system's \ac{com} expressed in $\mathcal{F}_B$ is retrieved by:
\begin{equation}\label{eq:t_a}
    \resizebox{0.9\columnwidth}{!}{ $\bm{\Gamma}_a=\begin{bmatrix}
        \Gamma_1\\\Gamma_2\\\Gamma_3
    \end{bmatrix}=\begin{bmatrix}
        \big(T_{2,7}-T_{1,6}+(T_{3,8}-T_{4,5})C_{\alpha}\big)W+\tau_{3,4,5,8}S_{\alpha}\\T_{3,4,5,8} C_{\alpha}L-T_{1,2,6,7}L\\ (T_{4,5}-T_{3,8})S_{\alpha}W+\tau_{3,4,5,8}C_{\alpha}+\tau_{1,2,6,7}
    \end{bmatrix}$}.
\end{equation}
Moreover, the tilting capabilities allow the generation of thrusts along the interaction axis $\bm{x}_B$. The body frame $\mathcal{F}_B$ actuation force vector\footref{fn:repeat} $\bm{F}_a \in \mathbb{R}^{3}$ is given by:
\begin{equation}\label{eq:f_a}
    \bm{F}_a=\begin{bmatrix}
        F_1\\0\\F_3
    \end{bmatrix}=\begin{bmatrix}
        T_{3,4,5,8}S_{\alpha}\\0\\T_{3,4,5,8}C_{\alpha}+T_{1,2,6,7}
    \end{bmatrix}.
\end{equation}
The system therefore has 5-\ac{dof} actuation provided by 9 system inputs including the rotors' rotating speed and back rotors' tilting angle. We denote $\bm{u}=\begin{bmatrix}
    \alpha&\Omega_1&\Omega_2&...&\Omega_8
\end{bmatrix}^{\top} \in \mathbb{R}^9$ as the system inputs.

\section{control framework}\label{s4:ctrl_des}
\begin{figure}[t]
   \centering
\includegraphics[width=\columnwidth]{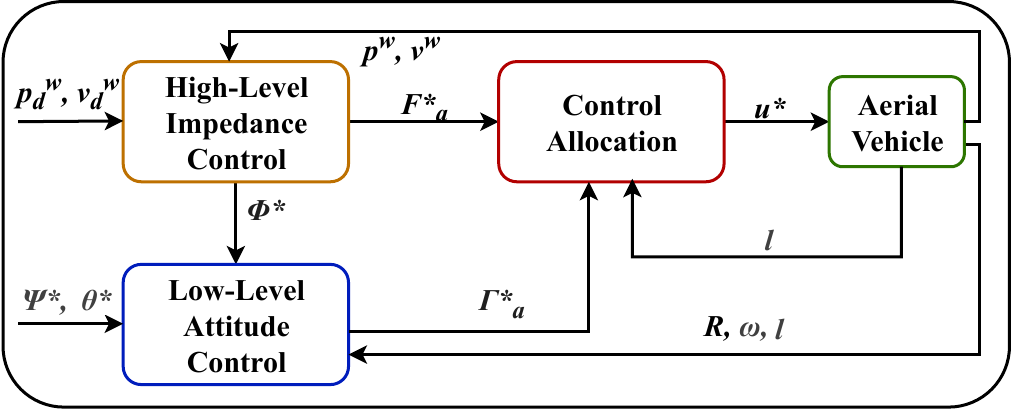}
    \caption{Overall control framework.}
    \label{fig:control}
    %\vspace{-0.5cm}
\end{figure}  
Considering the modeled system's under-actuation as introduced in \cref{s3:sys_model} and inspired by \cite{ding, bodie2019, lee2013}, we employ a cascade control pipeline as summarized in \cref{fig:control} 
with the capability to handle both free flights and physical interactions. In the following, $(\cdot)^*$ 
 indicate the desired quantities. 
This control framework integrates low-level geometric attitude control and high-level selective impedance control outputting the desired actuation wrenches $F_a^*$, $\Gamma_a^*\in\mathbb{R}^3$. These outputs are sent to the control allocation to compute the desired rotors' rotating speed $\Omega_i^*$ and tilting angle $\alpha^*$.
\subsection{Low-level Geometric Attitude Control}\label{sec:attitude_control}
We use the geometric controller in~\cite{lee2013} to control the system's attitude dynamics. The low-level controller outputs the desired actuation torques $\bm{\Gamma}_a^*$ by tracking the system's orientation and angular velocity in the body frame. The attitude tracking error $\bm{e}_R \in \mathbb{R}^3$ is defined as
$
    \bm{e}_R=\frac{1}{2}(\bm{R}^{*\top}\bm{R}-\bm{R}^{\top}\bm{R}^*)^{\vee},
$
and the angular velocity tracking error $\bm{e}_{\omega} \in \mathbb{R}^3$ is defined as:
$
    \bm{e}_{\omega}=\bm{\omega}-\bm{R}^{\top}\bm{R}^* \bm{\omega}^*
$. We define the integral error $\bm{e}_I=\int_o^t \bm{e}_{\omega}(\tau)+c_2\bm{e}_R(\tau)d\tau$, with $c_2$ being a positive constant.
A nonlinear attitude dynamics feedback controller is designed as the following:
\begin{equation}
    \begin{split}
    \bm{\Gamma}_a^*(l)=&-\bm{K}_R \bm{e}_R-\bm{K}_{\omega}\bm{e}_{\omega}-\bm{K}_I\bm{e}_I+\bm{\omega} \times \bm{I}(l)\bm{\omega}\\
    &+\bm{G}_{ang}-\bm{I}(l)(\mathbb{S}(\bm{\omega})\bm{R}^{\top}\bm{R}^*\bm{\omega}^*-\bm{R}^{\top}\bm{R}^*\dot{\bm{\omega}}^*),
\end{split}\label{eq:torque}
\end{equation}
where $\bm{K}_R$, $\bm{K}_{\omega}$, and $\bm{K}_I \in \mathbb{R}^{3\times3}$ are positive-definite matrices. 
\subsection{High-level Selective Impedance Control}\label{sec:impedance}
The high-level selective impedance control outputs the desired actuation forces $\bm{F}_a^*$ by tracking the body frame origin $O_B$ attached to the aerial vehicle's \ac{cog}, instead of the shifting-mass plate in the world frame. The system's linear dynamics can be re-written as:
\begin{equation}\label{eq:euqation_moton_w}
    \bm{M}\dot{\bm{\upsilon}}^w+\mathbb{S}(\bm{\omega}^w) \bm{M}\bm{\upsilon}^w+mg\bm{e}_3=\bm{F}_a^w+\bm{F}_C^w,
\end{equation}
where $\bm{F}_a^w=\bm{R}\bm{F}_a$ and $\bm{F}_C^w=\bm{R}\bm{F}_C$. %With ~\cref{eq:rotation_matrix}
Considering the frame transformation and  \cref{eq:f_a}, $\bm{F}_a^w$ is given by:
\begin{equation}\label{eq:f_a_w}
   \bm{F}_a^w=\begin{bmatrix}
        C_{\theta}C_{\psi}F_1+ (S_{\theta}C_{\psi}C_{\phi}+S_{\psi}S_{\phi})F_3 \\ C_{\theta}S_{\psi}F_1+(S_{\theta}S_{\psi}C_{\phi}-C_{\psi}S_{\phi})F_3\\
        -S_{\theta}F_1+ C_{\theta}C_{\phi}F_3
    \end{bmatrix}.
\end{equation}

The position tracking error is given by  
$
\bm{e}_p^w=\bm{p}^w-\bm{p}^{w*}
$, 
where $\bm{p}^w\in \mathbb{R}^3$ represents the position of the origin $O_B$ \ac{wrt} $O$ expressed in the world frame. The velocity tracking error can then be displayed as $\bm{e}_v^w=\bm{\upsilon}^w-\bm{\upsilon}^{w*}$ with $\bm{\upsilon}^{w*}=\dot{\bm{p}}^{w*}$. The desired close loop dynamics of the system with the selective impedance control is:
\begin{equation}\label{eq:close_loop}
    \bm{M}\dot{\bm{e}}_v^w+\bm{D}\bm{e}_v^w+\bm{K}\bm{e}_p^w=\bm{F}_C^w,
\end{equation}
where $\bm{D}$ and $\bm{K} \in \mathbb{R}^{3\times3}$ are positive-definite matrices representing the desired damping and stiffness of the system. 
Combining~\cref{eq:euqation_moton_w} and~\cref{eq:close_loop}, the desired actuation force vector expressed in the world frame is defined as:
\begin{equation}\label{eq:impedance}
\begin{split}
    \bm{F}_a^{w*}&=\begin{bmatrix}
       F_1^{w*}&F_2^{w*}&F_3^{w*}
\end{bmatrix}^{\top}\\
&=\bm{M}\dot{\bm{\upsilon}}^{w*}-\bm{D}\bm{e}_v^w-\bm{K}\bm{e}_p^w+\bm{\omega}^w\times \bm{M}\bm{\upsilon}^w+mg\bm{e}_3.
   \end{split}
\end{equation}

The system introduces coupling between its roll angle $\phi$ and its linear motion in the world frame not having a thrust component along the body axis $\bm{y}_B$. Again, combining \cref{eq:f_a_w} and \cref{eq:impedance} the desired roll angle $\phi^*$ is calculated via:
\begin{equation}\label{eq:desired_roll}
\begin{split}
    \phi^*=&atan2\big(S_{\psi}F_1^{w*}-C_{\psi}F_2^{w*},S_\theta C_\psi F_1^{w*}+...\\
    &S_\theta S_\psi F_2^{w*}+C_{\theta}F_3^{w*}\big).
    \end{split}
\end{equation}
 Furthermore, by using trigonometric calculations with~\cref{eq:f_a_w}, the desired actuation force vector $\bm{F}_a^*=\begin{bmatrix}
     F_1^*&0&F_3^*
 \end{bmatrix}^{\top}$ expressed in the body frame is given by:
 \begin{subequations}
 \begin{equation}
     F_1^*=C_\theta C_\psi F_1^{w*}+C_\theta S_\psi F_2^{w*}-S_\theta F_3^{w*},
     \end{equation}
     \begin{equation}
          \begin{split}
      F_3^*=&\big((S_{\psi}F_1^{w*}-C_{\psi}F_2^{w*})^2+(S_\theta C_\psi F_1^{w*}+...\\
      &S_\theta S_\psi F_2^{w*}+C_\theta F_3^{w*})^2\big)^{\frac{1}{2}}.
 \end{split}
     \end{equation}
 \end{subequations}
 The desired roll angle $\phi^*$ is then fed to the low-level attitude controller in \cref{sec:attitude_control} with the attitude references $\psi^*$ and $\theta^*$. 
 Finally, $\bm{F}_a^*$ and $\bm{\Gamma}_a^*$ are fed to the control allocation to derive the desired system inputs $\bm{u}^*$, as in \cref{fig:control}.
 \subsection{Control Allocation}
The control allocation problem for the examined case can be divided into finding the desired tilting angle $\alpha^*$ value,
\begin{equation}\label{eq:alpha}
    \alpha^*=\frac{\pi}{2}-atan2\big(F_3^*L+\Gamma_2^*(l),2LF_1^*\big),
\end{equation}
and retrieve the corresponding motor velocities $\Omega_i$. To handle the redundancy of the system, we introduce a virtual control input vector $\bm{\lambda}=\begin{bmatrix}
    \Omega_1^2 & \Omega_2^2 & ... & \Omega_8^2\end{bmatrix}^\top \in \mathbb{R}^{8}$. By applying~\cref{eq:shift_com},~\cref{eq:f_a}, and~\cref{eq:t_a} the desired virtual control vector is calculated via pseudo-inverse:
\begin{equation}\label{eq:lambda}
    \bm{\lambda}^*=\bm{H}(\alpha^*)^{\dag}\cdot  \begin{bmatrix}
    \bm{F}_a^*\\\bm{\Gamma}_a^*
    \end{bmatrix},
\end{equation}
where $\bm{H}(\alpha^*) \in \mathbb{R}^{6\times12}$ is the allocation matrix depending on the desired tilting angle $\alpha^*$ and $(\cdot)^{\dag}$ is the associate pseudo-inverse operator. Finally, we can compute the desired real system inputs as $\bm{u}^*=\begin{bmatrix} \alpha^* & \sqrt{\bm{\lambda}^*} \end{bmatrix}$. This approach minimizes the motors' rotating speed with guaranteed positive values~\cite{mina}.
%\import{}{texts/05_hardware}
\section{Software Integration}\label{s6:sftw}
\begin{figure}[t]
   \centering
\includegraphics[trim={0.6cm 0.2cm 0.6cm  0.2cm},clip,width=\columnwidth]{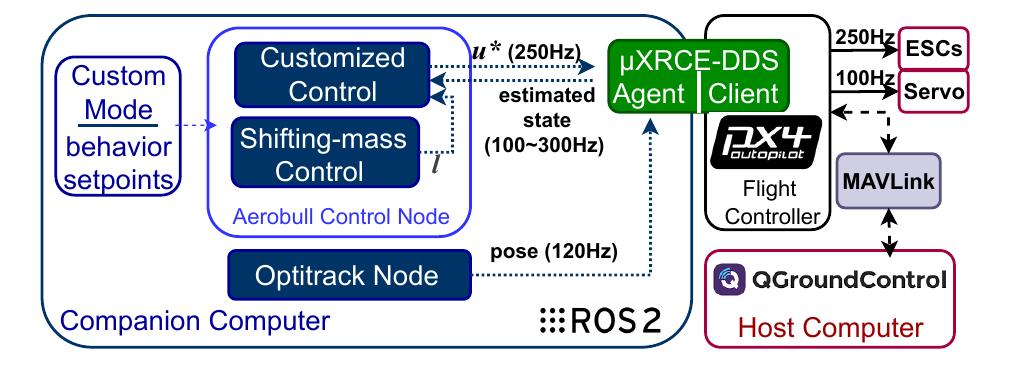}
    \caption{Software architecture scheme: the main connections are depicted.}
    \label{fig:software}
    %\vspace{-0.5cm}
\end{figure}

This section describes the software integration to implement the designed control framework on the developed hardware prototype in \cref{sec:hardware}. The proposed architecture in \cref{fig:software} is intended to establish a foundation for the future development of applications built on this platform creating a reusable framework that can be adapted for developing future custom aerial manipulation platforms.

The integration focuses on several key quality attributes according to the ISO/IEC $25010$\footnote{\url{https://iso25000.com/index.php/en/iso-25000-standards/iso-25010}} standard:
\begin{itemize}
    \item \textbf{Safety and Reliability}: Ensuring safety with fail-safes, pre-flight checks, and reliable software to minimize risks.
    \item \textbf{Real-time Performance}: Achieving stable flight through consistent timing and bounded scheduling latency.
    \item \textbf{Control Performance}: Supporting precise motion control with low steady-state errors, emphasizing stability over rapid trajectory tracking.
    \item \textbf{Extensibility}: Designing a modular architecture compatible with future custom platforms and tasks, using ROS2 for flexibility.
    \item \textbf{Maintainability}: Facilitating easy updates and debugging through a container-based development environment and diagnostic tools.
    \item \textbf{Usability}: Integrating with user-friendly interfaces like \texttt{QGroundControl} for effective prototyping.
\end{itemize}
\subsection{Detailed Software Architecture}
The developed software runs on the companion computer in ad-hoc Docker image using ROS2 and communicates with the flight controller via $\mu$XRCE-DDS protocol where PX4 \cite{px4} runs, as in~\cref{fig:software}. A key feature of the PX4 Autopilot is the flight mode which triggers different \ac{uav} behaviors. The host computer is only for flight-mode management and monitoring using \texttt{QGroundControl} which communicates with the flight controller through the MAVLink protocol. The ROS2 node - Aerobull Control Node, contains the customized control framework introduced in \cref{s4:ctrl_des} and the shifting-mass control which maps the servo position into the shifting-mass plate position.

Being different from the standards PX4 flight modes, this architecture includes a \textit{Custom Mode} based on the PX4 ROS2 Interface Library\footnote{\url{https://docs.px4.io/main/en/ros2/px4_ros2_interface_lib.html}\label{px4_note}}, a new ROS2-based C++ library expanding and simplifying the external interfacing capabilities of the PX4 Autopilot. This library provides a high-level \texttt{Control Interface} to facilitate the creation of customized control logic and a \texttt{Navigation Interface} for external localization sources integration. We set the \texttt{Navigation Interface} interconnecting the Optitrack system to the EKF (Extended Kalman Filter) state estimation module within PX4 using the relative ROS2 driver running in the companion computer.
\subsection{Control Interface}
The \texttt{Control Interface} enables the dynamic registration of flight modes in ROS2 mapped to the concept of \textit{robot behaviors}, such as taking off, landing, or navigation.
The \textit{Offboard Mode}\footref{px4_note} as one of the standard PX4 flight modes allows sending setpoints to PX4 while offering key advantages: it supports multiple on-demand applications triggered by events allowing custom modes with specific requirements and safety checks integrated into PX4’s arming and failsafe logic ensuring safe operations. The Aerobull \textit{Custom Mode} is defined starting from the \textit{Offboard Mode} enabling autonomous navigation and physical interaction using the customized controller in \cref{fig:control} by sending setpoints.

To do so, we extend the current PX4 ROS2 \texttt{Control Interface} capabilities to support the customized controller abstraction. 
% To avoid significant changes in the current library design, we propose to extend the \texttt{Setpoint} interface to represent the controller modules as \textit{high-level setpoints}. These setpoints interact with \textit{low-level setpoints}, for instance, the direct actuator commands. 
% Contrary to the regular setpoints, the controllers are active entities that can run a periodic timer callback to execute the control loop and update other setpoints, similar to the modes. As setpoints, they can also receive updates from a higher-level entity, either a mode or another controller. These updates correspond to the new controller reference to be processed by the respective control loop.
The developed \texttt{Control Interface} with direct actuator setpoints $\bm{u^*}$ as in \cref{fig:software} bypass the default PX4 cascaded control stack to be implemented in the ROS2 domain. This implementation allows abstaining completely from the PX4 architecture, serving as an alternative approach to ad-hoc firmware modifications~\cite{dangeloICUAS23, dangeloICUAS24}. Switching to the original flight stack using the RC is always possible preserving the original safety layers and controllers.

Considering that the shifting-mass dynamics are slower than the tilting angle dynamics and that the system control law varies with the shifting-mass plate position, the companion computer handles the whole-body control running each loop and sending commands at the same frequency as the native flight controller. The running frequencies are depicted in \cref{fig:software}.
Moreover, we leverage the Linux real-time kernel or \texttt{PREEMPT\_RT} patch to improve the reliability of the controller execution. Our goal is to ensure the operating system runs the ROS2 control loop timer every $4$ms ($250$Hz) with minimal scheduling latency, thereby preventing delays in the actuator commands sent to PX4.
\section{Experiments}\label{s7:case_st}
This experimental section tested two case studies of physical interaction tasks with and without displacing the system \ac{com} using the developed platform in \cref{fig:real_platform} for benchmarking. With the experiments, we aim to emphasize the proposed platform characteristics and efficient high-force generation capability. In our previous work \cite{ral_aerobull}, we achieved a pushing force of $15$\si{\newton} with the developed platform. In this article, we present the additional experiments during which we achieved $28$\si{\newton} of pushing force which is almost the same as the system gravity force ($30.5\si{\newton}$). The corresponding experimental video is available here: \url{https://youtu.be/vMauKy6UpIU}.
\begin{figure*}[t!]
    \centering
    \subfloat[\textbf{Case 1}]{\label{fig:reaction_case1}\includegraphics[trim={0.0cm 0.0cm 0.0cm  0.0cm},clip,width = 0.33\linewidth]{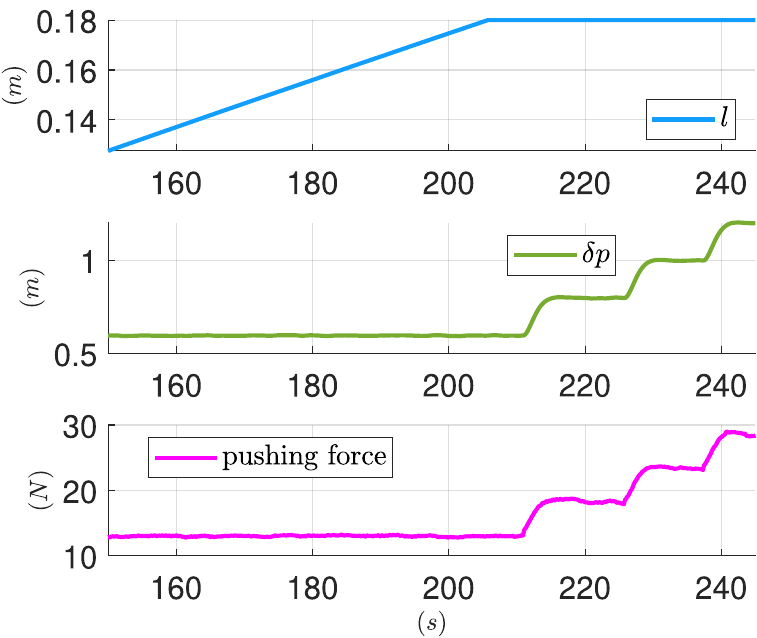}}
    \subfloat[\textbf{Case 1}]{\label{fig:rotor_case1}\includegraphics[trim={0.2cm 0.2cm 0.8cm  0.7cm},clip,width = 0.33\linewidth]{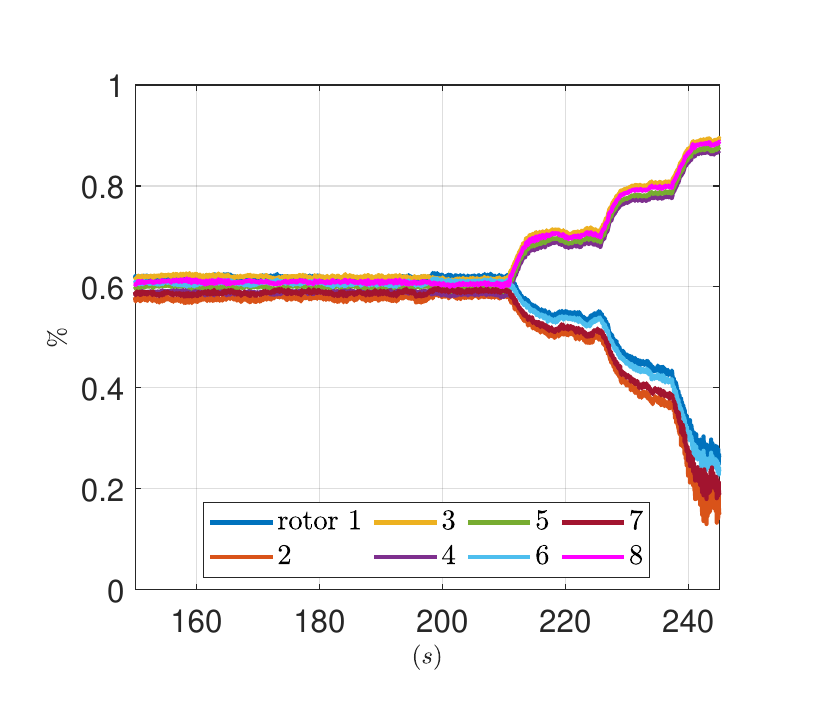}}
     \subfloat[\textbf{Case 1}]{\label{fig:att_case1}\includegraphics[trim={0.0cm 0.0cm 0.0cm 0.0cm},clip,width = 0.33\linewidth]{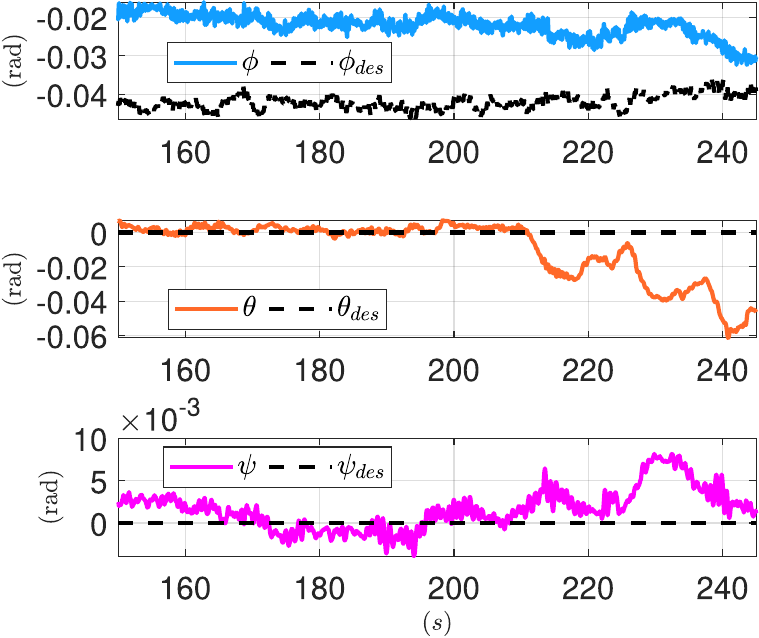}}\\
    \subfloat[\textbf{Case 2}]{\label{fig:reaction_case2}\includegraphics[trim={0.0cm 0.0cm 0.0cm  0.0cm},clip,width = 0.33\linewidth]{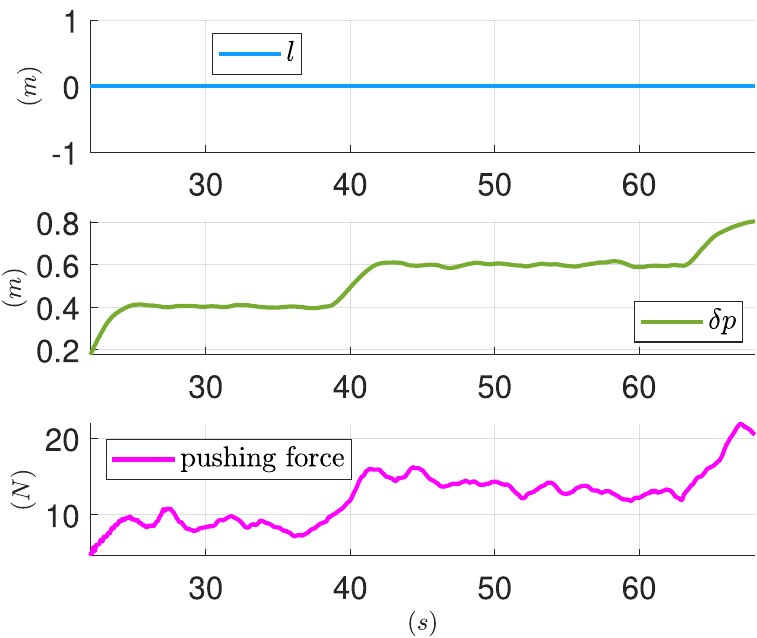}}
    \subfloat[\textbf{Case 2}]{\label{fig:rotor_case2}\includegraphics[trim={0.2cm 0.2cm 0.8cm  0.7cm},clip,width = 0.33\linewidth]{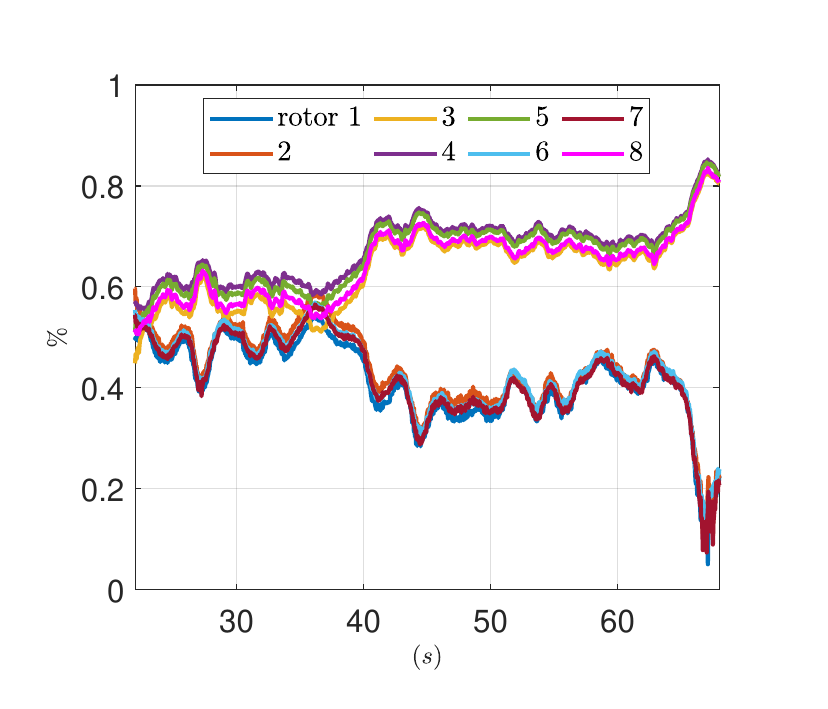}}
\subfloat[\textbf{Case 2}]{\label{fig:att_case2}\includegraphics[trim={0.0cm 0.0cm 0.0cm 0.0cm},clip,width = 0.33\linewidth]{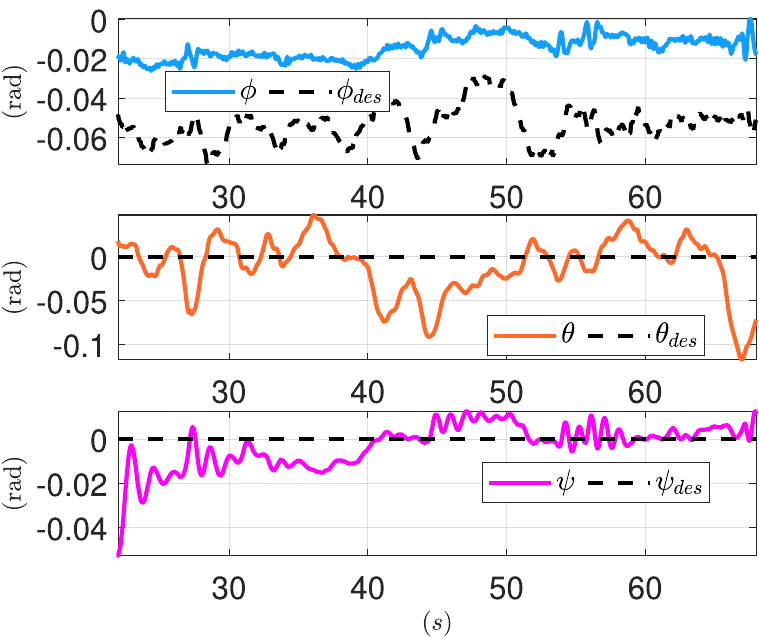}}
    \caption{Comparison between \textbf{Case 1} and \textbf{Case 2}: (a) and (d) from top to bottom represent the shifting-mass plate position $l$, setpoint relative location $\delta p$, and the exerted pushing force respectively; (b) and (e) illustrate the percentage of the rotor's rotating speed \ac{wrt} its maximum speed while (c) and (f) show the attitude tracking during interactions: dashed lines represent the reference attitude setpoints while continuous ones are the \ac{uav} feedback.}
\end{figure*}
\subsection{Experiment Setup}
\begin{table}[t!]
\centering
\caption{Control Parameters.}
\begin{tabularx}{\columnwidth}{p{0.2\textwidth}|X}
\toprule
\textbf{Gains} & \textbf{Value} \\ \hline
$\bm{K}$ & $diag([22.0, \ 22.0, \ 80.0])$ \\ 
$\bm{D}$ & $diag([10.0, \ 10.0, \ 45.0])$  \\ 
$\bm{K}_R$ & $diag([5.0, \ 5.0, \ 3.0])$\\ 
$\bm{K}_{\omega}$ & $diag([1.0, \ 1.4, \ 0.25])$ \\ 
$\bm{K}_I$ & $diag([0.0, \ 3.25, \ 0.5])$  \\
\hline
$c_2$ & $0.8$  \\ \hline
\end{tabularx}
\label{table_control}
\end{table}
The experiments were carried out in an indoor environment equipped with an OptiTrack system. An aluminum crane with a safety rope was used to avoid platform damage in case of a crash, and a vertical wooden board was mounted on the crane as the work surface (see~\cref{fig:action_1}). The OptiTrack system tracks the pose of the work surface and the \ac{uav}. 
The~\cref{table_control} provides the tuned controller gains while the aerodynamic coefficients of rotors are displayed in \cref{s2:sys_des}. In both cases, the platform was commanded to approach and push on the work surface with a series of position setpoints culminating with the ones \enquote{behind} the work surface. We denote $\delta p$ as the distance from the work surface to the setpoint along the interaction axis $\bm{x}_B$, and $\delta p>0$ for the ones \enquote{behind} the surface. In practice, the maximum shifting-mass plate position was configured such that $l_{max}=0.18\si{\meter}\ll\frac{m}{m_S}L=0.48\si{\meter}$ to avoid \ac{com} over-displacement, where $\frac{m}{m_S}L$ is the modeled maximum shifting-mass plate position in \cref{s2:sys_des}.

In \textbf{Case 1}, $\delta p=0.6$\si{\meter} was sent to the platform to establish an initial contact with the work surface. While preserving the stable contact with the work surface and the system orientation, the shifting-mass plate was shifted toward the work surface until its maximum position $0.18$\si{\meter}, as introduced in \cite{ral_aerobull}, where the system tracking capabilities are already demonstrated. 
With the displaced system \ac{com} resulted by moving the shifting-mass plate, position setpoints with $\delta p=0.8\si{\meter}, \ 1.0\si{\meter}, \ 1.2\si{\meter}$ were sent to the platform to exert ever-increasing pushing forces on the work surface. 

In \textbf{Case 2}, the platform was commanded to push on the work surface with $\delta p=0.4\si{\meter}, \ 0.6\si{\meter}, \ 0.8\si{\meter}, \ 1.0\si{\meter}$ while the shifting-mass plate remained still (i.e., fixed system \ac{com}). In the following, we present the experimental results of both cases.
\subsection{\textbf{Case 1}}
Concerning the previously achieved $15\si{\newton}$ of pushing force in~\cite{ral_aerobull}, in this case, the platform pushed on the work surface starting with a setpoint $0.6\si{\meter}$ \enquote{behind} the work surface. Without changing the setpoint location, the shifting-mass plate was sent forward along the body axis $\bm{x}_B$ while the system consistently exerted a pushing force of around $13\si{\newton}$, see \cref{fig:reaction_case1}. Different setpoints were sent to the high-level impedance controller after achieving the maximum shifting-mass plate position at $l=0.18\si{\meter}$. With $\delta p=0.8\si{\meter}$, the platform generated around $19\si{\newton}$ of pushing force on the work surface. With $\delta p=1.0\si{\meter}$, the pushing force reached $23\si{\newton}$, and finally the platform achieved a maximum pushing force of above $28\si{\newton}$ with $\delta p=1.2\si{\meter}$. The percentage of each rotor's rotating speed \ac{wrt} its saturation is displayed in \cref{fig:rotor_case1}. The back rotors' rotating speed increased with $\delta p$ reaching $0.9$ of the maximum speed at $\delta p=1.2\si{\meter}$, while the front rotors kept decreasing. While the shifting-mass plate was moving, the \ac{uav} maintained its orientation as shown in the corresponding attitude tracking in \cref{fig:att_case1}. The pitch and yaw error increased when a larger $\delta p$ was sent to the controller, as shown in the middle and bottom figures of \cref{fig:att_case1}.
\subsection{\textbf{Case 2}}
For benchmarking, we conducted additional pushing tasks using the same platform but without moving the shifting-mass plate (i.e., $l=0\si{\meter}$) with the same controller gains in~\cref{table_control}. Initially, a setpoint of $\delta p=0.4\si{\meter}$ was sent to the impedance controller, with which the platform established contact with the work surface and exerted a pushing force of around $10\si{\newton}$ with more oscillations compared to \textbf{Case 1}, as shown in \cref{fig:reaction_case2}, \cref{fig:rotor_case2}, and \cref{fig:att_case2}. Further, $\delta p$ was increased to $0.6\si{\meter}$, and the platform generated a pushing force of around $13\si{\newton}$ with explicit oscillations. Once $0.8\si{\meter}$ was sent to the controller, the system's oscillation amplified strongly reaching an impulse of $20\si{\newton}$ pushing force, and the experiment was terminated immediately after contacting to avoid platform damage. In this phase, the back rotors' highest rotating speed reached $0.85$ of the maximum rotating speed. While in \textbf{Case 1} with the same $\delta p$ value, the platform could stably exert a pushing force on the surface with the highest rotor saturation level of $0.62$. 
Later, an additional setpoint of $\delta p=1.0\si{\meter}$ was directly sent to the system to establish initial contact. The system ran into instability after contacting the surface: these data are not illustrated in the figures but the experiment is shown in the accompanying video. 
\subsection{Discussion}
The comparison between these two cases validated the high-force generation capability of the designed system with displaced \ac{com}. With such a system design featuring the tiltable back rotors and a displaceable \ac{com}, the platform with a total mass of $3.12\si{\kilogram}$ can exert a pushing force of $28\si{\newton}$ that is almost the same as its gravity force. The designed system showed less oscillation and higher force generation during a pushing task on a vertical surface when its \ac{com} was shifted toward the work surface as in \cref{fig:actions}.
\section{Conclusion}\label{s8:end}
With this paper, we extended the results achieved in~\cite{ral_aerobull} where a novel aerial manipulation platform is presented. This platform features a displaceable \ac{com} and tiltable back rotors, tailored to advanced aerial manipulation tasks on non-horizontal surfaces. In this paper, we focused on the system design, hardware and software integration, and the platform's performance during the task execution in terms of force generation. The benefit of \ac{com} displacement for efficient high-force interaction is highlighted in the experiments compared to the platform with a fixed \ac{com} configuration. We identified promising future revenue by presenting a novel solution to enhance high-force tool manipulation with aerial vehicles in industrial applications.

\vspace{-33pt}

\begin{IEEEbiography}[{\includegraphics[width=1in,height=1.25in, clip,keepaspectratio]{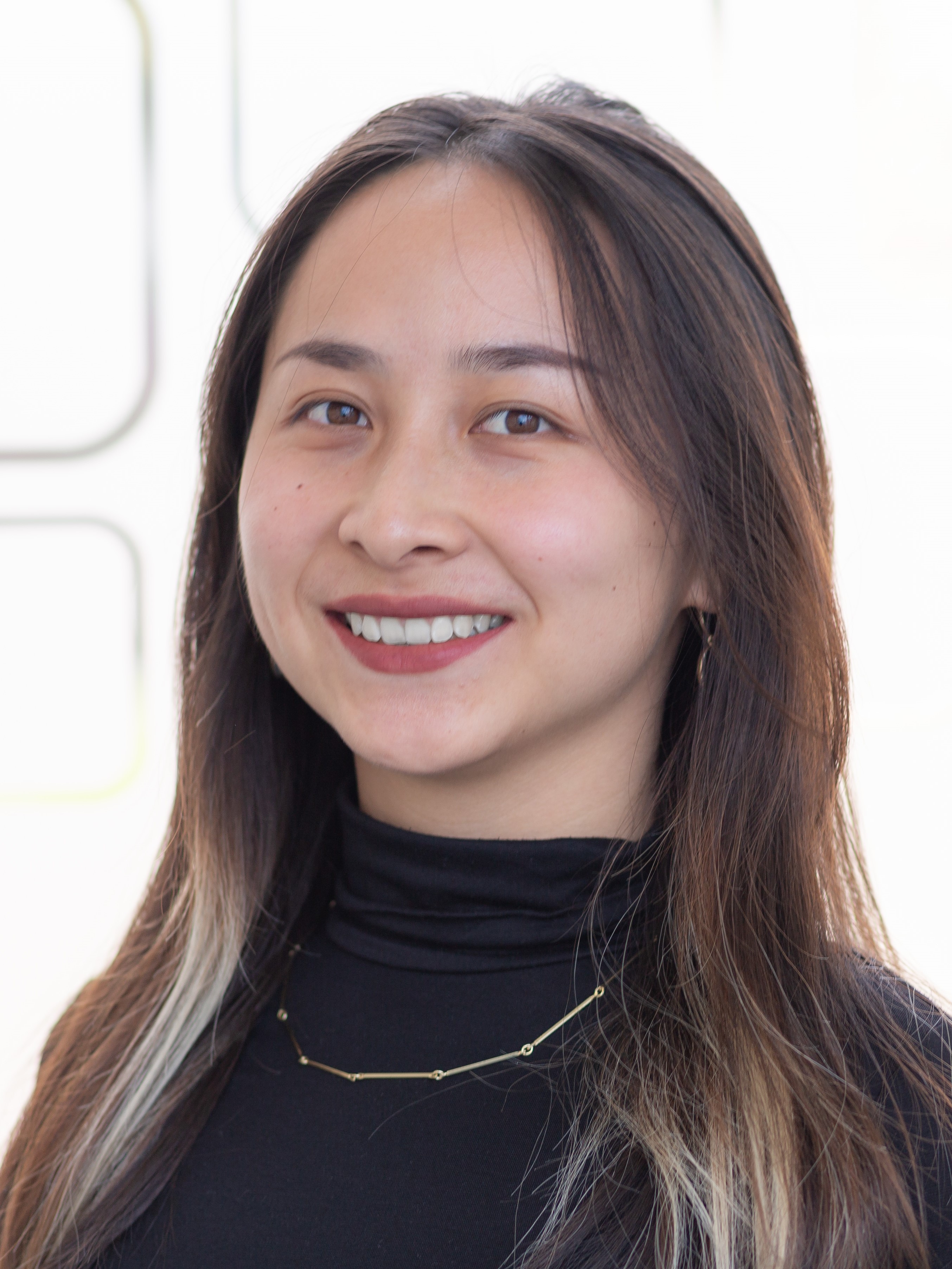}}]{Tong Hui} (Student Member, IEEE) received the B.S. double-degree in mechanical engineering from Tongji University (Shanghai, China) and Politecnico di Milano (Milan, Italy), in 2018 and 2019 respectively. She received her M.S. in mechanical engineering from Politecnico di Milano in 2021. She worked as a research fellow at the Italian Institute of Technology afterward. She is currently a PhD candidate at the Technical University of Denmark with the Department of Electrical and Photonics Engineering. Her current research interests include robotics, aerial manipulation, physical interactions, robotic system design, and control.
\end{IEEEbiography}
\begin{IEEEbiography}
[{\includegraphics[trim={2.0cm 12.5cm 2.0cm 0.0cm}, clip,width=1in,height=1.25in, clip,keepaspectratio]{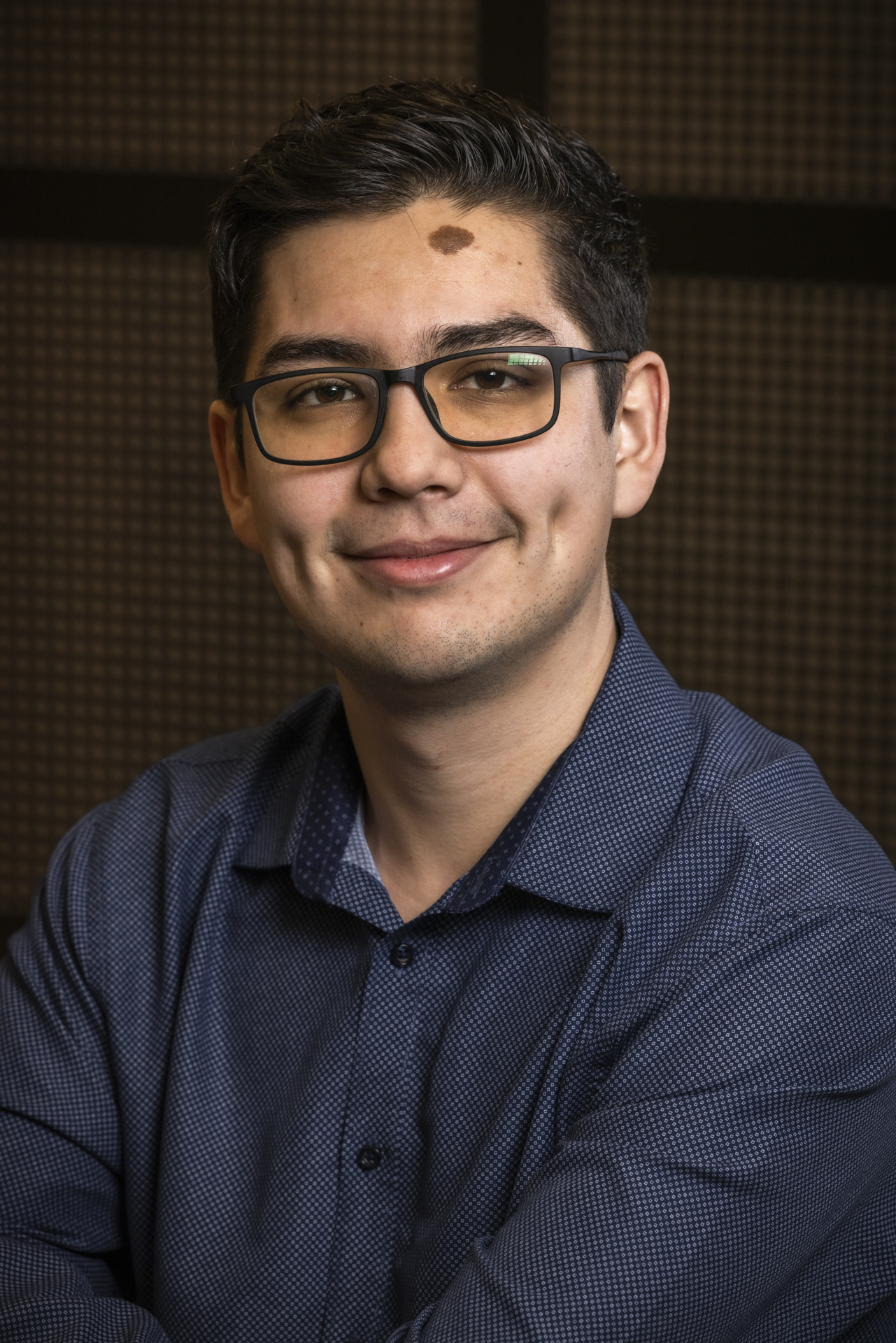}}]{Esteban Zamora} received the B.S. degree in electrical engineering from the University of Costa Rica, in 2018. He is currently a MSc student in Autonomous Systems from the Technical University of Denmark and a Robotics Software Engineer at AgriRobot Aps. His research interests include perception and control in autonomous systems, aerial robotics, agricultural robotics, and embedded systems.
\end{IEEEbiography}
\begin{IEEEbiography}
[{\includegraphics[width=1in,height=1.25in, clip,keepaspectratio]{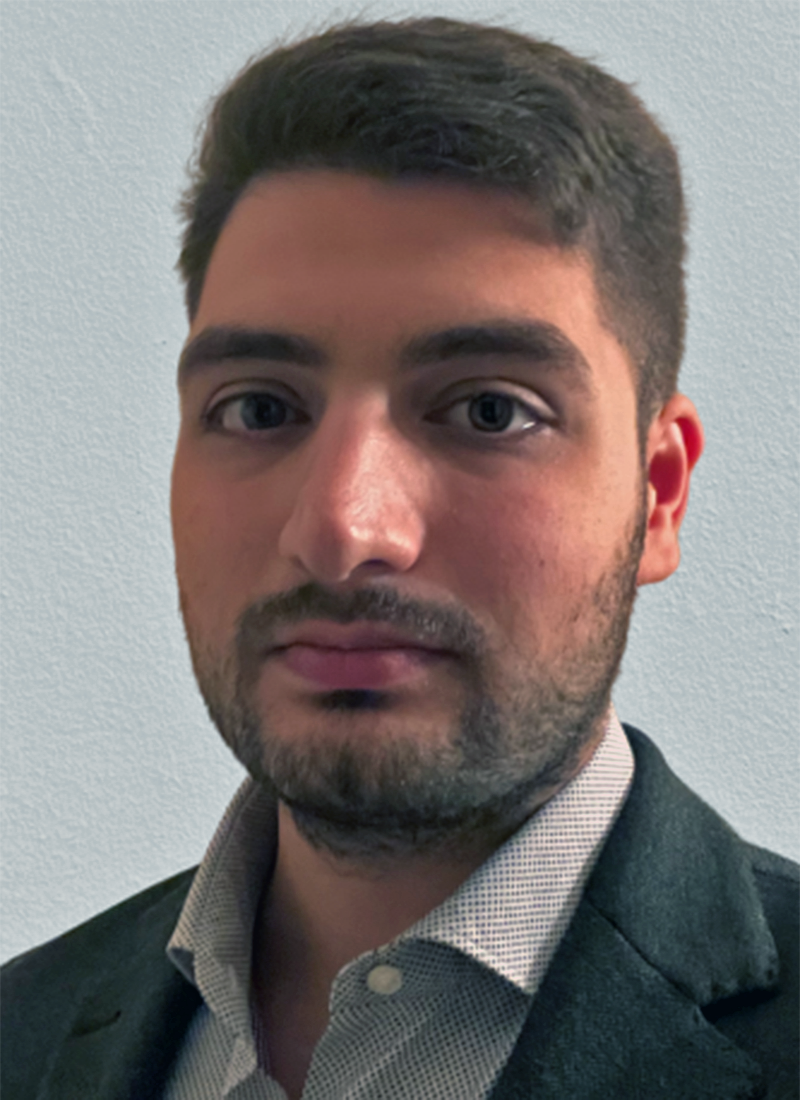}}]{Simone D'Angelo} (Student Member, IEEE) received the Bachelor's and Master’s degree in Automation Engineering from the University of Naples Federico II in 2018 and 2021, respectively. He is currently a Ph.D. candidate at the Department of Electrical Engineering and Information Technology at the University of Naples Federico II and part of the PRISMALab aerial robotics research group. His current research interests include autonomous systems, robotics, mechatronics, aerial manipulation, physical interaction, and hybrid force/vision control.
\end{IEEEbiography}
\begin{IEEEbiography}
[{\includegraphics[trim={0.2cm 0.0cm 0.8cm  0.0cm},width=1in,height=1.25in, clip,keepaspectratio]{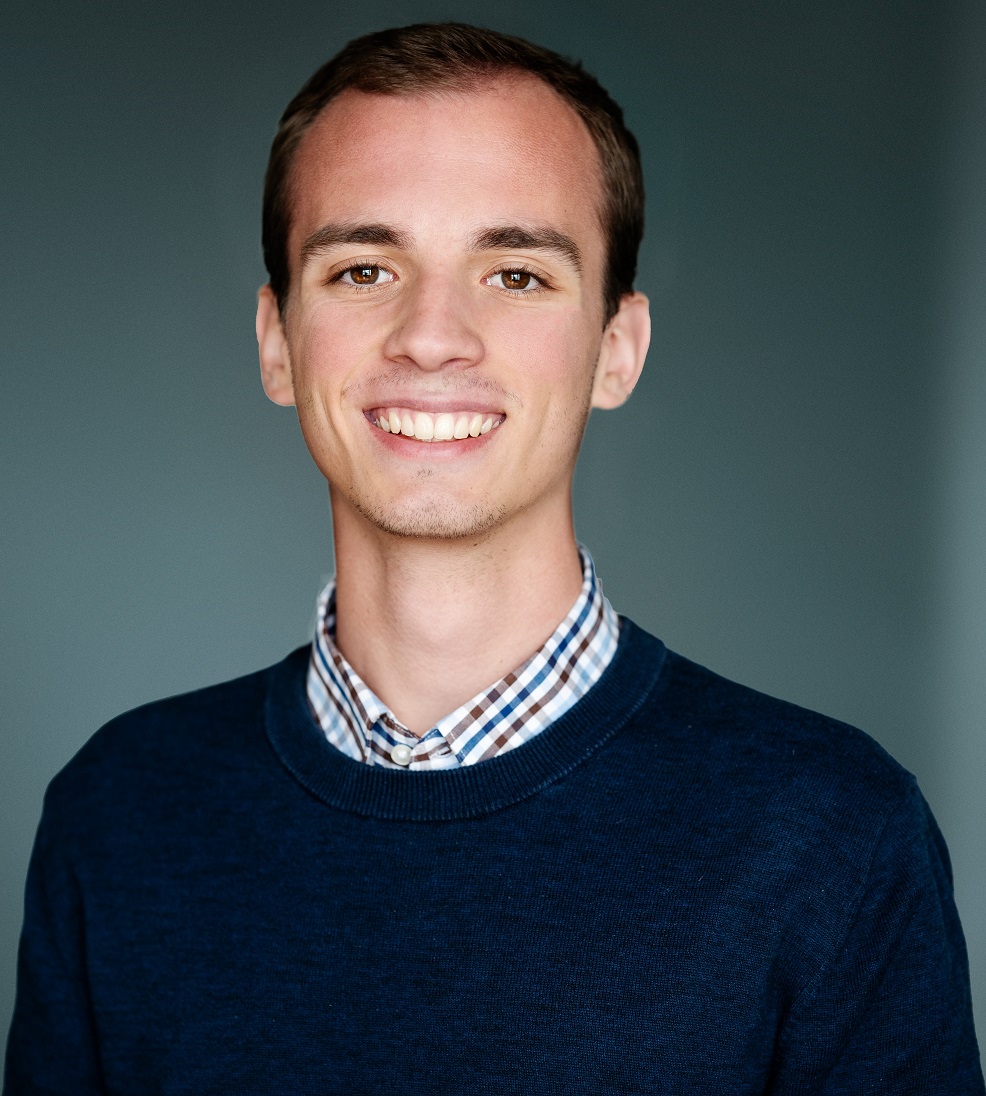}}]{Stefan Rucareanu} is currently completing a double degree between Ecole Centrale de Nantes in France and the Technical University of Denmark (DTU) for the title of robotics engineer of Centrale Nantes and an MSc in Autonomous Systems engineering at DTU. His research interests are focused on autonomous systems, aerial robotics, mechanics, and control.
\end{IEEEbiography}
\begin{IEEEbiography}
[{\includegraphics[width=1in,height=1.25in,clip,keepaspectratio]{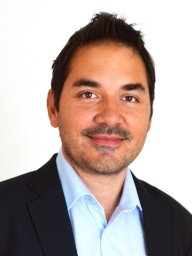}}]{Matteo Fumagalli}  (Member, IEEE) is an Associate Professor in Robotics and Unmanned Autonomous Systems at the Technical University of Denmark, a position he has held since April 2019. He began his academic career in 2015 at Aalborg University (DK) as an Assistant Professor in Mechatronics, following a postdoctoral research role in aerial robotics at the Robotics and Mechatronics group of the University of Twente (NL, 2011-2015). Dr. Fumagalli earned his PhD in Humanoid Technologies from the University of Genoa in 2011, conducting research on force estimation for humanoid robots during his tenure at the Italian Institute of Technology. He also holds an MSc (2006) and a BSc (2003) in Mechanical Engineering from the Polytechnic University of Milan (Italy).
His research centers on the physical interaction of floating-based systems and the mechatronic development of innovative physically interacting robotic bodies.
\end{IEEEbiography}
\vspace{11pt}

\vfill

\end{document}